\documentclass{article} 
\usepackage{iclr2027_conference,times}

\usepackage{amsmath,amsfonts,bm}

\def\eqref#1{equation~\ref{#1}}

\def\1{\bm{1}}

\DeclareMathAlphabet{\mathsfit}{\encodingdefault}{\sfdefault}{m}{sl}
\SetMathAlphabet{\mathsfit}{bold}{\encodingdefault}{\sfdefault}{bx}{n}

\usepackage{hyperref}
\usepackage{url}
\usepackage{booktabs}
\usepackage{amssymb}
\usepackage{graphicx}
\graphicspath{{figures/}{./}}
\usepackage{tikz}
\usetikzlibrary{arrows.meta,positioning,calc,shapes.geometric,shapes.misc,
                decorations.pathreplacing,decorations.pathmorphing,
                decorations.markings,patterns,fadings,backgrounds,fit,
                matrix,intersections,math}

\definecolor{inkgray}  {HTML}{2F3437}   
\definecolor{softgray} {HTML}{EDEEF0}   
\definecolor{eegblue}  {HTML}{3D5A98}   
\definecolor{eeglight} {HTML}{DCE4F5}
\definecolor{vizteal}  {HTML}{1B7F81}   
\definecolor{vizlight} {HTML}{D3E9E9}
\definecolor{latred}   {HTML}{C9483F}   
\definecolor{latlight} {HTML}{FADFDC}
\definecolor{bislate}  {HTML}{4A7FBF}   
\definecolor{bilight}  {HTML}{E2ECF9}
\definecolor{actgold}  {HTML}{E0A02E}   
\definecolor{actlight} {HTML}{FBEBCB}
\definecolor{losspur}  {HTML}{7A5AA6}   
\definecolor{losslight}{HTML}{EBE3F5}

\tikzset{
  every node/.append style={font=\small, text=inkgray},
  box/.style      ={draw=inkgray, line width=0.7pt, rounded corners=2pt,
                    align=center, inner xsep=4.5pt, inner ysep=4pt,
                    font=\small},
  mod/.style      ={box, draw=eegblue, fill=eeglight},          
  vmod/.style     ={box, draw=vizteal, fill=vizlight},          
  frozen/.style   ={box, draw=inkgray!55, fill=softgray},       
  actbox/.style   ={box, draw=actgold!85!black, fill=actlight}, 
  lossbox/.style  ={box, draw=losspur, fill=losslight, rounded corners=4pt},
  latbox/.style   ={box, draw=latred, fill=latlight},
  bibox/.style    ={box, draw=bislate, fill=bilight},
  panel/.style    ={draw=inkgray!25, line width=0.6pt, rounded corners=3pt,
                    fill=softgray!45, inner sep=5pt},
  tag/.style      ={font=\scriptsize\itshape, text=inkgray!75, inner sep=1.5pt},
  lbl/.style      ={font=\scriptsize, text=inkgray!85, inner sep=2pt},
  flow/.style     ={-{Stealth[length=2.1mm,width=1.5mm]}, line width=0.85pt,
                    draw=inkgray!85, rounded corners=2pt},
  vflow/.style    ={flow, draw=vizteal},
  eflow/.style    ={flow, draw=eegblue},
  aflow/.style    ={flow, draw=actgold!85!black, line width=1.0pt},
  ghost/.style    ={-{Stealth[length=1.9mm,width=1.4mm]}, line width=0.7pt,
                    draw=inkgray!45, dashed, dash pattern=on 1.6pt off 1.3pt},
  lossflow/.style ={ghost, draw=losspur!80},
  sgmark/.style   ={font=\scriptsize, text=losspur, inner sep=1.5pt},
}

\newcommand{\snowflake}[1]{%
  \begin{tikzpicture}[baseline=-0.6ex,scale=#1]
    \foreach \a in {0,60,120}{\draw[line width=0.45pt,inkgray!70]
      (\a:0.9ex)--(\a+180:0.9ex);}
    \foreach \a in {0,60,120,180,240,300}{%
      \draw[line width=0.4pt,inkgray!70] (\a:0.55ex)--($(\a:0.55ex)+(\a+35:0.3ex)$);
      \draw[line width=0.4pt,inkgray!70] (\a:0.55ex)--($(\a:0.55ex)+(\a-35:0.3ex)$);}
  \end{tikzpicture}}
\newcommand{\frostlabel}[1]{#1\,\snowflake{1}}

\newcommand{\eegtrace}[4]{%
  \begin{scope}
    \foreach \c [count=\i from 0] in {1,...,#3}{%
      \pgfmathsetmacro{\yo}{-\i*(#2)/(#3)}
      \draw[line width=0.45pt, draw=#4, opacity=0.9]
        plot[domain=0:#1, samples=90, smooth]
        (\x, {\yo + 0.030*(#2)*sin(360*(2.1+0.7*\i)*(\x/(#1)))
                  + 0.022*(#2)*sin(360*(5.3+1.3*\i)*(\x/(#1))+40*\i)
                  + 0.014*(#2)*sin(360*(11.0+2.0*\i)*(\x/(#1))+90*\i)});}
  \end{scope}}

\newcommand{\headtop}[3]{%
  \begin{scope}
    \fill[#2, opacity=0.55] (0,0) -- (90:#1) arc (90:270:#1) -- cycle;
    \fill[#3, opacity=0.55] (0,0) -- (90:#1) arc (90:-90:#1) -- cycle;
    \draw[line width=0.6pt, inkgray!80] (0,0) circle (#1);
    \draw[line width=0.6pt, inkgray!80]
      (100:#1) -- ++(80:0.22*#1) coordinate (nosetip) -- (80:#1);
    \draw[line width=0.5pt, inkgray!45, dash pattern=on 1.2pt off 1.2pt]
      (0,-#1) -- (0,#1);
  \end{scope}}

\AddToHook{env/table/begin}{%
  \setlength{\abovecaptionskip}{0pt}%
  \setlength{\belowcaptionskip}{5pt}%
}

\title{Mind World Models: Joint Learning of Motor EEG Representations and Action-Conditioned World Dynamics}

\author{Weikang Qiu\\
Yale University
\And
Zheng Huang \\
Dartmouth College
\And
Tinglin Huang \\
Yale University
\And
Yujun Yan \\
Dartmouth College
\And
Rex Ying \\
Yale University
}

\newcommand{\method}{MWM}

\newcommand{\resultClsTasks}{9}
\newcommand{\resultRegTasks}{7}
\newcommand{\resultClsGainPct}{5.9}
\newcommand{\resultRegGainPct}{16.2}

\newcommand{\resultWMWithinGapX}{3.3}
\newcommand{\resultWMSubjectGapX}{4.7}

\newif\ifarxiv
\arxivtrue
\ifarxiv\iclrfinalcopy\fi
\begin{document}

\maketitle
\ifarxiv\lhead{}\fi

\begin{abstract}
Motor decoding from electroencephalography (EEG) supports non-invasive brain--computer interfaces, yet existing methods either rely on costly cue-based trials and coarse categorical labels or use pretraining objectives that do not directly capture how movement unfolds, limiting their ability to learn transferable motor representations. To this end, we propose the Mind World Model (MWM), a framework that learns motor EEG representations from scalable naturalistic EEG and egocentric video without action annotations. MWM treats EEG observed up to the current frame as the latent action of a world model: a mind model encodes the EEG, and a world model uses it to predict the representation of a future frame, yielding a movement-grounded pretraining objective. We further introduce the Lateralization-Aware Transformer (LAT), which uses frame averaging to separate shared bilateral from side-specific lateral features, encoding hemispheric correspondence to reduce the burden of learning left--right relationships from data. MWM pretraining produces two models: a mind model for downstream motor decoding and a world model that can be adapted to prediction conditioned on real actions, opening a scalable and flexible path to collect egocentric data for embodied AI.
Across \resultClsTasks{} classification and \resultRegTasks{} regression tasks, the mind model improves task-averaged balanced accuracy by \resultClsGainPct{}\% and Pearson correlation by \resultRegGainPct{}\% relative to the strongest baseline in each task family. Compared with an inverse-dynamics latent-action baseline, the world model achieves lower prediction error and benefits \resultWMWithinGapX{}$\times$ more from matched real actions within recordings and \resultWMSubjectGapX{}$\times$ more on held-out subjects.

\end{abstract}

\section{Introduction}

\suppressfloats[t]
\begin{figure}[t]
\centering
\input{figures/mwm_teaser}
\caption{\textbf{Existing approaches (top) versus \method{} (bottom) on the three challenges of motor EEG representation learning.}
\textbf{(1)} Cue-based trials are costly to scale, and categorical labels give one class to movements that unfold differently~\citep{tangermann2012review,schirrmeister2017deep,lawhern2018eegnet}.
\textbf{(2)} General-purpose EEG foundation models (FMs) pretrain with EEG-only objectives~\citep{kostas2021bendr,jiang2024large,wang2025cbramod}.
\method{} answers both with one mechanism: it treats the EEG recorded up to the current frame as the latent action of a world model that predicts the next visual state. It therefore learns from naturalistic EEG and egocentric video without action annotations, and its target is the visual consequence of behavior rather than the EEG itself.
\textbf{(3)} Existing methods do not explicitly separate bilateral from lateral activity~\citep{schirrmeister2017deep,lawhern2018eegnet,wang2025cbramod}; \method{} does, and pairs each horizontal video mirror with an exchange of only the lateral component $\mathbf{H}^{\mathrm{lat}}$.}
\label{fig:teaser}
\end{figure}

Motor decoding from electroencephalography (EEG) aims to infer intended or executed movements from non-invasive recordings of brain activity. It encompasses both recognizing movement categories and estimating continuous quantities such as movement trajectories and force. This capability underpins brain--computer interfaces for assistive control and neurorehabilitation, allowing movement-related neural activity to guide external devices or rehabilitation feedback~\citep{tangermann2012review,ramos2013brain}. For these applications, useful EEG representations must capture both which movement is intended and how it unfolds, and transfer across motor tasks with limited labeled data. Learning such representations presents three challenges (Figure~\ref{fig:teaser}, top).

First, task-specific motor decoding faces limitations in both data scalability and the information conveyed by supervision. Supervised motor-decoding research is largely organized around labeled examples of predefined tasks, commonly collected through repeated, cue-based laboratory trials~\citep{tangermann2012review,schirrmeister2017deep,lawhern2018eegnet}. Such protocols make it costly to expand the volume and diversity of training data. Moreover, categorical annotations such as ``left hand'' and ``right hand'' compress motor behavior into a small vocabulary. Movements with different trajectories, speeds, or forces can share the same label, leaving their within-class variation unspecified by the training target. Direct measurements of continuous kinematics or force retain such within-class variation, but collecting them at scale typically requires task-specific instrumentation and careful synchronization with EEG. Thus, scaling motor representation learning requires both more accessible collection paradigm and supervision that retains information beyond discrete action categories.

Second, general-purpose EEG foundation models address label scarcity through self-supervised objectives such as contrastive learning, masked signal reconstruction, and neural token prediction~\citep{kostas2021bendr,chien2022maeeg,jiang2024large,wang2024eegpt,wang2025cbramod}. Although these objectives learn useful signal structure, they do not directly supervise the relationship between neural activity and movement. Their pretraining corpora are also assembled primarily for broad or clinical EEG analysis rather than for capturing continuous, naturalistic motor behavior and its consequences~\citep{jiang2024large,wang2025cbramod,obeid2016tuh,ouahidi2025reve}. These differences motivate pretraining that aligns both the learning signal and the recording context with motor decoding.

Third, motor EEG exhibits \emph{lateralization}: similar movement-related patterns can indicate different acting sides depending on which hemisphere they occur in. For example, sensorimotor rhythm suppression is often stronger over the left hemisphere for right-hand movements and vice versa~\citep{pfurtscheller1997motor,nam2011lateralization}. Without this structural prior, existing encoders~\citep{schirrmeister2017deep,lawhern2018eegnet,wang2025cbramod} must learn the relationship between patterns and hemispheric location from data, potentially requiring larger training datasets and yielding suboptimal decoding performance when data are limited.

World-model training faces a similar bottleneck in scaling data collection. Instrumented interfaces can constrain natural behavior and capture only selected aspects of action. For example, handheld grippers and motion-capture gloves primarily record hand kinematics, without directly measuring head and leg movements, interaction forces, or hand stiffness~\citep{chi2024umi,wang2024dexcap}. Video-based approaches instead infer pseudo-actions from current and future frames~\citep{bruce2024genie,gao2025adaworld}. However, access to future frames introduces a risk of target-information leakage: pseudo-actions may encode future visual content, allowing the world model to exploit prediction shortcuts rather than learn transferable action-conditioned dynamics~\citep{garrido2026lawm}.

To address these challenges, we propose the Mind World Model (\method{}), which jointly learns a mind model and a world model from synchronized EEG and egocentric video in EgoBrain~\citep{lin2026egobrain}. The mind model maps EEG observed up to the current frame to a latent action, and the world model uses it to predict a future visual representation. Using egocentric recordings of natural motor behavior, rather than general-purpose EEG corpora, aligns the training data with motor decoding. Visual prediction supplies supervision beyond categorical labels without task-specific annotation, enabling scalable pretraining. Encoding actions from EEG avoids explicit motion measurements and future-frame leakage. Joint learning thus produces action-related EEG representations and a world-model initialization for future prediction conditioned on real actions.

To capture motor lateralization, we introduce the Lateralization-Aware Transformer (LAT), an EEG encoder that uses frame averaging~\citep{puny2022frame} over original and mirrored representations to separate shared bilateral features from side-specific lateral features. When a video is mirrored horizontally, LAT exchanges lateral EEG features across homologous electrodes while preserving the bilateral component. This prior links corresponding movement patterns across hemispheres, reducing the burden of learning their left--right relationship from limited data.

Our contributions are as follows:
\begin{itemize}
    \item We introduce \method{}, which learns motor EEG representations from naturalistic EEG--video by predicting the visual consequences of behavior. This objective provides supervision beyond categorical action labels and enables scalable data collection without explicit action annotations.
    \item We propose LAT to encode hemispheric correspondence by separating bilateral and lateral EEG features, reducing the burden of learning this structure from data.
    \item We show that the pretrained world model transfers from EEG-derived actions to prediction conditioned on real actions, opening a path to scalable world-model data collection without relying on manipulation interfaces that constrain natural behavior and capture only selected aspects of action.
\end{itemize}

We evaluate the mind model on \resultClsTasks{} classification and \resultRegTasks{} regression tasks, and the world model on future prediction conditioned on real actions. Relative to the strongest baseline in each task family, the mind model improves task-averaged balanced accuracy by \resultClsGainPct{}\% and Pearson correlation by \resultRegGainPct{}\%. Ablations show that alignment, prediction, and frame averaging each improve average transfer, with effects that vary by task. Relative to the latent-action baseline, the world model achieves lower prediction error and benefits \resultWMWithinGapX{}$\times$ more from matched real actions within recordings and \resultWMSubjectGapX{}$\times$ more on held-out subjects; it also outperforms an identically trained action-free predictor.

\section{Related Works}

\noindent\textbf{EEG Foundation Models.}
EEG foundation models learn transferable representations through contrastive learning, masked reconstruction, and token prediction~\citep{kostas2021bendr,chien2022maeeg,jiang2024large,wang2024eegpt}. Such objectives exploit structure in unlabeled recordings, reducing dependence on task-specific annotations and enabling transfer across EEG analysis tasks. Other work improves spatiotemporal modeling and electrode-layout flexibility~\citep{wang2025cbramod,wang2025eegmamba,doner2025luna}, scales pretraining~\citep{ouahidi2025reve}, or incorporates language and complementary neural modalities~\citep{jiang2025neurolm,xiao2025brainomni}. These advances target general EEG representation learning; \method{} uses visual consequences of natural behavior to supervise motor representations and explicitly models hemispheric correspondence.

\noindent\textbf{World Models.}
World models support prediction and control through observation generation~\citep{ha2018recurrent,micheli2023transformers,alonso2024diffusion} or compact latent dynamics~\citep{hafner2020dream,hansen2024tdmpc2}. Recent approaches predict frozen or jointly learned visual features~\citep{zhou2025dinowm,sobal2025rewardfree,assran2025vjepa2,maes2026leworldmodel}. Predicting in feature space lets the dynamics model focus on abstract state changes without reconstructing every detail of the observation. Video-based methods infer latent actions from frame transitions~\citep{bruce2024genie,gao2025adaworld,garrido2026lawm}. However, collecting explicit action labels can be difficult to scale, while inferring actions from future frames risks leaking target information into the action representation~\citep{garrido2026lawm}. \method{} pretrains a world model from EEG and video, deriving actions without explicit motion measurements or future frames. The pretrained model transfers to prediction conditioned on real actions.

\section{Method}

Our central design is to use neural activity in place of an externally recorded action within a predictive world model. The mind model observes EEG only up to the current visual state and must encode the information needed to predict what happens next. We first formulate this prediction task and its two model components, then describe the Lateralization-Aware Transformer (LAT) and training objectives.

\begin{figure}[t]
\centering
\input{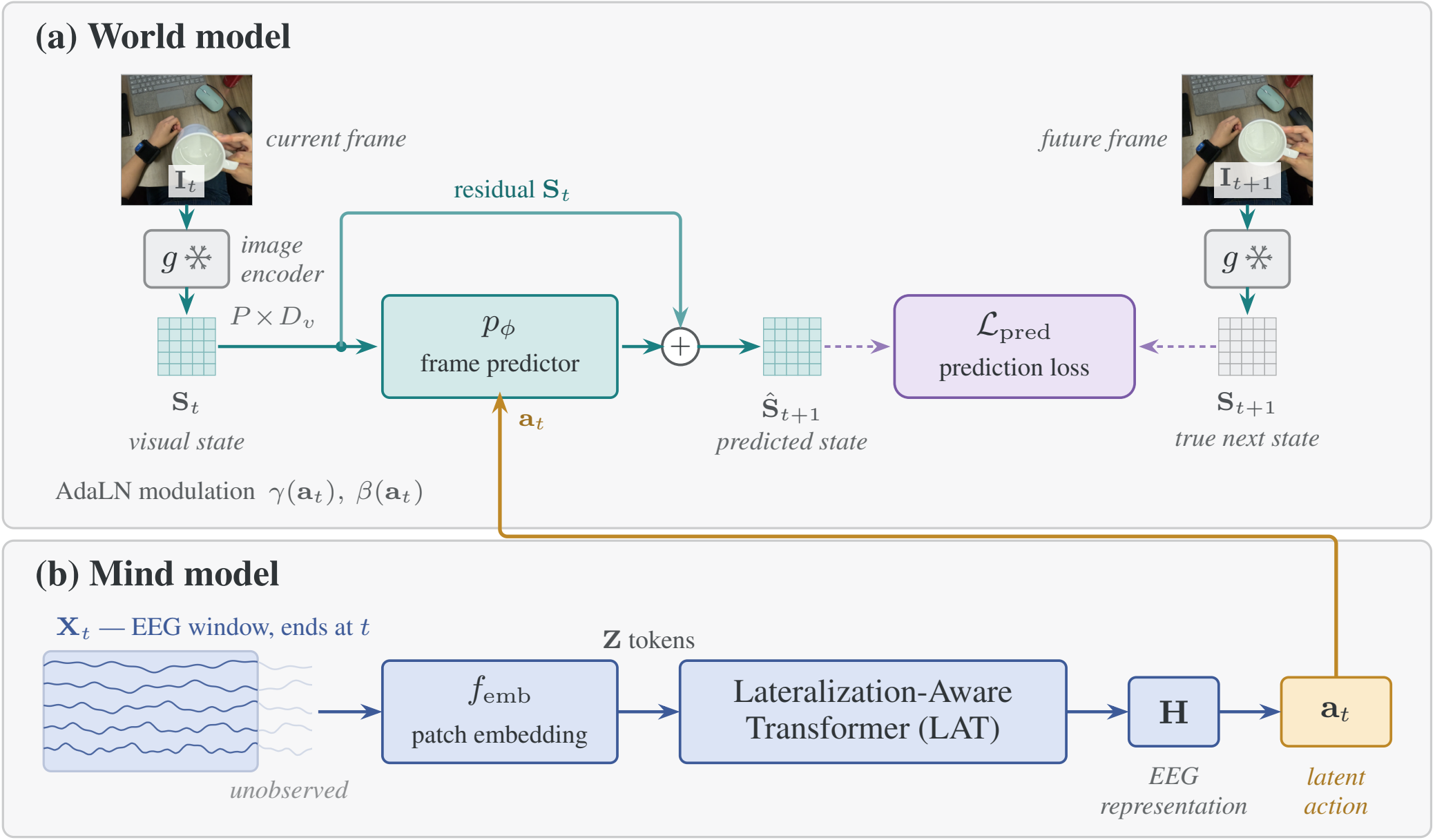}
\caption{\textbf{The Mind World Model.} \textbf{(a)} A frozen encoder $g$ maps the current
frame to patch features that a residual predictor advances; the same frozen $g$ builds
the target from the next frame. \textbf{(b)} The mind model encodes the EEG window, which
ends on the current frame; its LAT encoder is detailed in
Figure~\ref{fig:mirrorarch}. The EEG reaches the predictor only as AdaLN modulation, never as an
input token: it occupies the slot a recorded action would.}
\label{fig:framework}
\end{figure}

\subsection{World Model}

As illustrated in Figure~\ref{fig:framework}, the Mind World Model (\method{}) jointly learns motor-related EEG representations and visual dynamics from synchronized EEG and egocentric video without action annotations. Given an EEG segment $\mathbf{X}_t\in\mathbb{R}^{C\times T}$ and a current RGB frame $\mathbf{I}_t\in\mathbb{R}^{3\times H_v\times W_v}$, the mind model uses LAT (Section~\ref{sec:mind-model}) to encode the EEG into a latent action $\mathbf{a}_t\in\mathbb{R}^{D_a}$, which conditions the world model's prediction of the future visual state. Here, $C$ is the number of EEG channels, $T$ the number of time samples, $H_v$ and $W_v$ the image height and width, and $D_a$ the action dimension.

\noindent\textbf{Image Encoder.}
As shown in Figure~\ref{fig:framework}(a), the current frame $\mathbf{I}_t$ and its future frame $\mathbf{I}_{t+1}\in\mathbb{R}^{3\times H_v\times W_v}$ are independently encoded by a frozen DINOv2 encoder $g$~\citep{oquab2024dinov2}:
\begin{equation}
    \mathbf{S}_t=g(\mathbf{I}_t)\in\mathbb{R}^{P\times D_v}, \qquad
    \mathbf{S}_{t+1}=g(\mathbf{I}_{t+1})\in\mathbb{R}^{P\times D_v},
\end{equation}
where $P$ is the number of visual patches and $D_v$ is their feature dimension. We use DINOv2 to balance performance and computational efficiency. Freezing $g$ provides stable targets and reduces training compute and memory by eliminating encoder backpropagation.

\noindent\textbf{Frame Predictor.}
The frame predictor $p_{\phi}$ advances the current visual state using the EEG-derived latent action:
\begin{equation}
    \hat{\mathbf{S}}_{t+1}
    =
    p_{\phi}(\mathbf{S}_t,\mathbf{a}_t)\in\mathbb{R}^{P\times D_v}.
\end{equation}
The mind-model pathway in Figure~\ref{fig:framework}(b) aggregates EEG tokens into $\mathbf{a}_t$, which enters the frame predictor only through modulation. The current-frame tokens are then processed by a lightweight Transformer whose layers are modulated by $\mathbf{a}_t$ through adaptive layer normalization (AdaLN)~\citep{peebles2023scalable}. For a hidden token $\mathbf{h}\in\mathbb{R}^{D_p}$, where $D_p$ is the predictor width, the action produces a scale $\gamma(\mathbf{a}_t)\in\mathbb{R}^{D_p}$ and a shift $\beta(\mathbf{a}_t)\in\mathbb{R}^{D_p}$:
\begin{equation}
    \operatorname{AdaLN}(\mathbf{h},\mathbf{a}_t)
    =
    \gamma(\mathbf{a}_t)\odot
    \operatorname{LN}(\mathbf{h})
    +
    \beta(\mathbf{a}_t).
\end{equation}
This design treats EEG as a latent action that controls the predicted visual transition. The residual connection in Figure~\ref{fig:framework}(a) adds a predicted change tensor $\Delta_{\phi}(\mathbf{S}_t,\mathbf{a}_t)\in\mathbb{R}^{P\times D_v}$ to the current state:
\begin{equation}
    \hat{\mathbf{S}}_{t+1}
    =
    \mathbf{S}_t+\Delta_{\phi}(\mathbf{S}_t,\mathbf{a}_t),
\end{equation}
such that the predictor focuses on dynamic changes instead of reconstructing the static scene.

To avoid temporal leakage, the EEG window ends at the current frame, while the prediction target is sampled from the subsequent time interval. Therefore, the target visual change is not already contained in the EEG observation window.

\subsection{Mind Model}
\label{sec:mind-model}

\begin{figure}[t]
\centering
\input{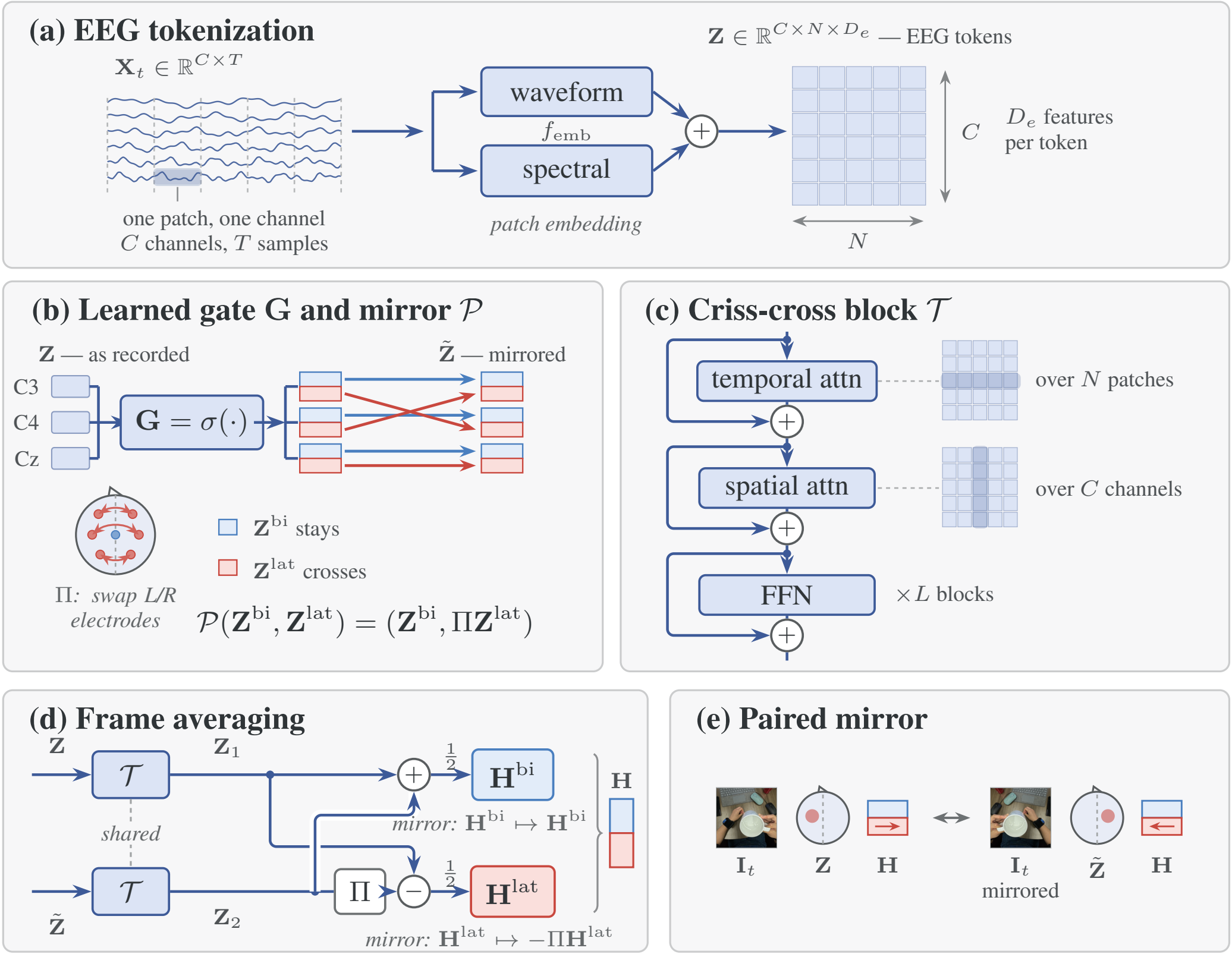}
\caption{\textbf{The Lateralization-Aware Transformer (LAT) for EEG encoding.} \textbf{(a)} Patch embedding, applied within each channel,
\textbf{(b)} the learned gate $\mathbf{G}\in[0,1]^{C\times N\times D_e}$ and the electrode permutation $\Pi\in\{0,1\}^{C\times C}$, which act
on the embedded tokens before any Transformer block; the mirror acts on the decomposed pair and is an involution, \textbf{(c)} a criss-cross block,
\textbf{(d)} a shared Transformer with invariant and equivariant readout heads; under the mirror, $\mathbf{H}^{\mathrm{bi}}$ is unchanged and $\mathbf{H}^{\mathrm{lat}}$ flips sign after aligning homologous channels, and
\textbf{(e)} the paired EEG--video mirror. Only the learned lateral component is exchanged
across hemispheres; the raw signal is never permuted.}
\label{fig:mirrorarch}
\end{figure}

\noindent\textbf{Transformers.}
Figure~\ref{fig:mirrorarch}(a) shows how each channel of $\mathbf{X}_t$ is divided into non-overlapping temporal patches. A channel-wise CNN embeds each patch by combining its temporal waveform and spectral representation, producing
\begin{equation}
    \mathbf{Z}_t
    =
    f_{\mathrm{emb}}(\mathbf{X}_t)
    \in\mathbb{R}^{C\times N\times D_e},
\end{equation}
where $N$ is the number of temporal patches per channel and $D_e$ is the EEG token width.

The criss-cross Transformer block in Figure~\ref{fig:mirrorarch}(c) alternates temporal attention across patches with spatial attention across EEG channels. This factorization preserves the channel--time structure of EEG while reducing the cost of full attention. The resulting patch tokens are aggregated into the latent action $\mathbf{a}_t$ used by the frame predictor.

\noindent\textbf{LAT and Paired Mirror Augmentation.}
LAT encodes the lateralization prior described in the introduction: corresponding movement patterns over opposite hemispheres indicate opposite acting sides. Modeling this correspondence explicitly allows LAT to learn shared movement structure across left--right counterparts.

Raw EEG mixes bilateral activity with lateralized motor activity; swapping entire channels would exchange both, although only the latter should change sides under our motor prior. We therefore learn the decomposition in feature space, as shown in Figure~\ref{fig:mirrorarch}(b). Let $\Pi\in\{0,1\}^{C\times C}$ exchange homologous electrodes, such as C3 and C4, along the channel axis, and write $\mathbf{Z}=\mathbf{Z}_t\in\mathbb{R}^{C\times N\times D_e}$. An elementwise gate $\mathbf{G}\in[0,1]^{C\times N\times D_e}$ separates bilateral and lateral components $\mathbf{Z}^{\mathrm{bi}},\mathbf{Z}^{\mathrm{lat}}\in\mathbb{R}^{C\times N\times D_e}$:
\begin{equation}
    \mathbf{Z}^{\mathrm{lat}}
    =
    \mathbf{G}\odot\mathbf{Z},
    \qquad
    \mathbf{Z}^{\mathrm{bi}}
    =
    (1-\mathbf{G})\odot\mathbf{Z}.
\end{equation}
We define the mirror operation as
\begin{equation}
    \mathcal{P}(\mathbf{Z}^{\mathrm{bi}},\mathbf{Z}^{\mathrm{lat}})
    =
    (\mathbf{Z}^{\mathrm{bi}},\,\Pi\mathbf{Z}^{\mathrm{lat}}).
\end{equation}
Thus, bilateral activity is preserved, while lateral activity is exchanged between the two hemispheres. Since the decomposition is computed once, $\mathcal{P}\circ\mathcal{P}=\mathrm{id}$ for any $\mathbf{G}$, as $\Pi^2=I$. We write $\tilde{\mathbf{Z}}=\mathbf{Z}^{\mathrm{bi}}+\Pi\mathbf{Z}^{\mathrm{lat}}$ for the recombined mirrored input.

Bilateral features capture movement structure shared by both sides and should remain invariant under mirroring, while lateral features encode the acting side and should transform equivariantly. Inspired by frame averaging (FA)~\citep{puny2022frame}, we wrap a shared Transformer with invariant and equivariant readouts (Figure~\ref{fig:mirrorarch}(d)). The same Transformer $\mathcal{T}$ encodes the original and mirrored inputs into $\mathbf{Z}_1,\mathbf{Z}_2\in\mathbb{R}^{C\times N\times D_e}$:
\begin{equation}
    \mathbf{Z}_1=\mathcal{T}(\mathbf{Z}), \qquad
    \mathbf{Z}_2=\mathcal{T}(\tilde{\mathbf{Z}}).
\end{equation}
The invariant head averages these outputs; the equivariant head takes their signed difference after aligning homologous channels. Each head uses a linear feature readout $r_{\mathrm{inv}},r_{\mathrm{eq}}:\mathbb{R}^{D_e}\to\mathbb{R}^{D_e/2}$, applied at each channel and time location, to produce $\mathbf{H}^{\mathrm{bi}},\mathbf{H}^{\mathrm{lat}}\in\mathbb{R}^{C\times N\times D_e/2}$:
\begin{equation}
    \mathbf{H}^{\mathrm{bi}}
    = r_{\mathrm{inv}}\!\left(\frac{\mathbf{Z}_1+\mathbf{Z}_2}{2}\right), \qquad
    \mathbf{H}^{\mathrm{lat}}
    = r_{\mathrm{eq}}\!\left(\frac{\mathbf{Z}_1-\Pi\mathbf{Z}_2}{2}\right).
\end{equation}
Mirroring the input, i.e., applying $\mathcal{P}$ to the decomposed pair, swaps the two passes ($\mathbf{Z}_1\leftrightarrow\mathbf{Z}_2$). Since $\Pi$ commutes with the channel-wise readouts, the outputs transform as
\begin{equation}
    \mathbf{H}^{\mathrm{bi}}\mapsto\mathbf{H}^{\mathrm{bi}}, \qquad
    \mathbf{H}^{\mathrm{lat}}\mapsto-\Pi\mathbf{H}^{\mathrm{lat}},
\end{equation}
so the bilateral output is exactly invariant and the lateral output flips sign after aligning homologous channels. Concatenating them yields the final EEG representation $\mathbf{H}=\left[\mathbf{H}^{\mathrm{bi}};\mathbf{H}^{\mathrm{lat}}\right]\in\mathbb{R}^{C\times N\times D_e}$.

Egocentric recordings can overrepresent dominant-hand movements because people preferentially use one hand, with right-handedness more common in the population~\citep{papadatou2020handedness}. To address this sampling bias, we randomly mirror each video horizontally during training and apply the corresponding operation $\mathcal{P}$ to its EEG representation, as illustrated in Figure~\ref{fig:mirrorarch}(e). This produces paired left--right counterparts while preserving their cross-modal correspondence.

\subsection{Training Objective}

\noindent\textbf{Alignment Loss.}
We align global EEG and synchronized-frame representations using symmetric InfoNCE~\citep{oord2018representation}, denoted $\mathcal{L}_{\mathrm{align}}$. EEG features are projected into the visual feature space, and both modalities are $\ell_2$-normalized before computing similarities.

\noindent\textbf{Prediction Loss.}
The prediction loss in Figure~\ref{fig:framework}(a) compares the predicted state with the frozen representation of the future frame. For each sample, we average the $\ell_1$ distance over the $P$ patches:
\begin{equation}
    \mathcal{L}_{\mathrm{pred}}
    =
    \frac{1}{P}
    \sum_{p=1}^{P}
    \left\|
    \hat{\mathbf{S}}_{t+1}^{p}
    -
    \mathbf{S}_{t+1}^{p}
    \right\|_1.
\end{equation}
Because the DINOv2 encoder $g$ is frozen, the visual target remains fixed throughout training. Minimizing this loss therefore requires the mind model to extract EEG information that explains the upcoming visual transition.

The overall training objective combines the standard masked EEG patch reconstruction loss $\mathcal{L}_{\mathrm{mask}}$ with visual alignment and prediction, weighted by nonnegative coefficients $\lambda_{\mathrm{align}}$ and $\lambda_{\mathrm{pred}}$:
\begin{equation}
    \mathcal{L}
    =
    \mathcal{L}_{\mathrm{mask}}
    +
    \lambda_{\mathrm{align}}\mathcal{L}_{\mathrm{align}}
    +
    \lambda_{\mathrm{pred}}\mathcal{L}_{\mathrm{pred}}.
\end{equation}

\section{Experiments}

We first evaluate the pretrained mind model on motor regression and classification benchmarks. We then analyze lateralization in its representations through hand-motion readouts and motor-imagery topographies, and use ablations to assess the contributions of the lateralization-aware architecture, prediction, and alignment. Finally, we evaluate the pretrained world model's transfer from EEG-derived actions to future prediction conditioned on real actions.

\subsection{Experimental Setup}

\noindent\textbf{Pretraining.}
We pretrain \method{} on 63.28 hours of synchronized EEG and egocentric video from all 40 EgoBrain participants~\citep{lin2026egobrain}. All regression and classification benchmarks use the same fixed pretrained checkpoint. Dataset preprocessing details are provided in Appendix~\ref{app:dataset-details}.

\noindent\textbf{Linear-probe evaluation.}
We evaluate every backbone on the same task-specific one-second EEG input sampled at 200\,Hz, freezing the backbone and training only a linear prediction head. We report test balanced accuracy for classification and test Pearson correlation for regression, averaging correlations across output dimensions for vector targets. Each table entry reports the mean and standard deviation over three evaluation seeds shared across models for that task. The participant-disjoint splits and evaluation settings are specified in Appendices~\ref{app:dataset-details} and~\ref{app:implementation-details}.

\noindent\textbf{Baselines.}
We compare with recent EEG foundation models: BIOT~\citep{yang2023biot}, LaBraM~\citep{jiang2024large}, CBraMod~\citep{wang2025cbramod}, CSBrain~\citep{zhou2025csbrain}, and REVE~\citep{ouahidi2025reve}.

\subsection{Regression Tasks}

\noindent\textbf{Tasks.}
We evaluate seven continuous decoding tasks in Table~\ref{tab:reg-pooled-lp}. ForceGameReg~\citep{li2026forcegame} predicts plateau force; SrisReachME~\citep{srisrisawang2024simultaneous}, OfnerApertureME, OfnerKinME~\citep{ofner2017upper}, and BiomechMag~\citep{perezblanco2026wrist} predict future reaching speed, hand-aperture change, hand speed, and wrist-speed magnitude, respectively. WAYGALReg~\citep{luciw2014multichannel} predicts grip and load forces, while ForenzoFutVec~\citep{forenzo2024continuous} predicts two-dimensional cursor velocity in a closed-loop motor-imagery interface. Dataset statistics, target definitions, preprocessing, and participant-disjoint splits are detailed in Appendix~\ref{app:dataset-details}.

\begin{table}[t]
\centering
\small
\setlength{\tabcolsep}{4pt}
\caption{Pearson correlation on regression tasks. Mean $\pm$ sample s.d. over three seeds.}
\label{tab:reg-pooled-lp}
\resizebox{\linewidth}{!}{%
\begin{tabular}{lccccccc}
\toprule
Model & ForceGameReg & SrisReachME & OfnerApertureME & OfnerKinME & WAYGALReg & BiomechMag & ForenzoFutVec \\
\midrule
BIOT & $0.044{\scriptstyle\pm0.025}$ & $0.102{\scriptstyle\pm0.014}$ & $0.032{\scriptstyle\pm0.008}$ & $-0.021{\scriptstyle\pm0.003}$ & $0.217{\scriptstyle\pm0.001}$ & $0.039{\scriptstyle\pm0.001}$ & $0.041{\scriptstyle\pm0.003}$ \\
LaBraM & $0.112{\scriptstyle\pm0.003}$ & $0.173{\scriptstyle\pm0.002}$ & $0.059{\scriptstyle\pm0.007}$ & $0.001{\scriptstyle\pm0.002}$ & $0.226{\scriptstyle\pm0.001}$ & $0.119{\scriptstyle\pm0.001}$ & $0.094{\scriptstyle\pm0.002}$ \\
CBraMod & $0.083{\scriptstyle\pm0.027}$ & $\underline{0.217{\scriptstyle\pm0.001}}$ & $0.075{\scriptstyle\pm0.007}$ & $0.078{\scriptstyle\pm0.011}$ & $0.324{\scriptstyle\pm0.003}$ & $0.138{\scriptstyle\pm0.001}$ & $0.096{\scriptstyle\pm0.006}$ \\
CSBrain & $\underline{0.113{\scriptstyle\pm0.003}}$ & $0.185{\scriptstyle\pm0.001}$ & $0.053{\scriptstyle\pm0.006}$ & $0.062{\scriptstyle\pm0.011}$ & $0.295{\scriptstyle\pm0.001}$ & $0.136{\scriptstyle\pm0.002}$ & $\underline{0.114{\scriptstyle\pm0.002}}$ \\
REVE & $0.089{\scriptstyle\pm0.021}$ & $\underline{0.217{\scriptstyle\pm0.001}}$ & $\underline{0.153{\scriptstyle\pm0.015}}$ & $\underline{0.175{\scriptstyle\pm0.010}}$ & $\mathbf{0.460{\scriptstyle\pm0.003}}$ & $\underline{0.153{\scriptstyle\pm0.005}}$ & $0.097{\scriptstyle\pm0.002}$ \\
\midrule
\textbf{MWM (ours)} & $\mathbf{0.165{\scriptstyle\pm0.010}}$ & $\mathbf{0.241{\scriptstyle\pm0.002}}$ & $\mathbf{0.233{\scriptstyle\pm0.018}}$ & $\mathbf{0.206{\scriptstyle\pm0.006}}$ & $\underline{0.433{\scriptstyle\pm0.002}}$ & $\mathbf{0.157{\scriptstyle\pm0.002}}$ & $\mathbf{0.127{\scriptstyle\pm0.001}}$ \\
\bottomrule
\end{tabular}}
\end{table}

\noindent\textbf{Results.}
In Table~\ref{tab:reg-pooled-lp}, \method{} improves task-averaged Pearson correlation by \resultRegGainPct{}\% relative to REVE, the strongest baseline by this measure. Gains span force, kinematic, and cursor-velocity targets. WAYGALReg is the exception, with REVE performing best. This task involves grip and load forces under varying object weight and surface friction~\citep{luciw2014multichannel}. We hypothesize that visual prediction provides a less direct supervisory signal for these force variations, which are not uniquely determined by visible movement.

\subsection{Classification Tasks}

\noindent\textbf{Tasks.}
We evaluate nine classification tasks in Table~\ref{tab:cls-lp}: PhysioNet-MI~\citep{schalk2009eegmmidb}, Schwarz~\citep{schwarz2020analyzing}, Kaya5F~\citep{kaya2018large}, Forenzo2024~\citep{forenzo2024continuous}, FineMI~\citep{yi2025multimodal}, Ofner2017ME and Ofner2017MI~\citep{ofner2017upper}, Weibo2014~\citep{yi2014evaluation}, and Jeong2020~\citep{jeong2020multimodal}. They cover motor execution, motor imagery, pre-movement grasp decoding, and cursor-control direction. Schwarz evaluates grasp type from EEG recorded 1--2\,s before movement onset. Dataset statistics, target definitions, preprocessing, and participant-disjoint splits are detailed in Appendix~\ref{app:dataset-details}.

\begin{table}[t]
\centering
\small
\setlength{\tabcolsep}{4pt}
\caption{Balanced accuracy on classification tasks. Mean $\pm$ sample s.d. over three seeds.}
\label{tab:cls-lp}
\resizebox{\linewidth}{!}{%
\begin{tabular}{lccccccccc}
\toprule
Model & PhysioNet-MI & Schwarz & Kaya5F & Forenzo2024 & FineMI & Ofner2017ME & Weibo2014 & Ofner2017MI & Jeong2020 \\
\midrule
BIOT & $0.262{\scriptstyle\pm0.008}$ & $0.507{\scriptstyle\pm0.007}$ & $0.207{\scriptstyle\pm0.008}$ & $0.272{\scriptstyle\pm0.004}$ & $0.172{\scriptstyle\pm0.009}$ & $0.149{\scriptstyle\pm0.020}$ & $0.161{\scriptstyle\pm0.003}$ & $0.147{\scriptstyle\pm0.005}$ & $0.097{\scriptstyle\pm0.003}$ \\
LaBraM & $0.470{\scriptstyle\pm0.006}$ & $0.522{\scriptstyle\pm0.018}$ & $0.231{\scriptstyle\pm0.007}$ & $0.297{\scriptstyle\pm0.002}$ & $0.179{\scriptstyle\pm0.014}$ & $0.160{\scriptstyle\pm0.008}$ & $0.185{\scriptstyle\pm0.020}$ & $\underline{0.158{\scriptstyle\pm0.001}}$ & $0.118{\scriptstyle\pm0.008}$ \\
CBraMod & $\underline{0.581{\scriptstyle\pm0.007}}$ & $0.519{\scriptstyle\pm0.003}$ & $\underline{0.266{\scriptstyle\pm0.008}}$ & $\underline{0.313{\scriptstyle\pm0.003}}$ & $\underline{0.199{\scriptstyle\pm0.002}}$ & $0.177{\scriptstyle\pm0.026}$ & $0.183{\scriptstyle\pm0.002}$ & $0.147{\scriptstyle\pm0.003}$ & $0.146{\scriptstyle\pm0.000}$ \\
CSBrain & $0.549{\scriptstyle\pm0.003}$ & $\underline{0.551{\scriptstyle\pm0.025}}$ & $0.244{\scriptstyle\pm0.005}$ & $0.303{\scriptstyle\pm0.001}$ & $0.183{\scriptstyle\pm0.002}$ & $0.174{\scriptstyle\pm0.008}$ & $0.162{\scriptstyle\pm0.001}$ & $0.148{\scriptstyle\pm0.005}$ & $\underline{0.149{\scriptstyle\pm0.006}}$ \\
REVE & $0.560{\scriptstyle\pm0.006}$ & $0.529{\scriptstyle\pm0.012}$ & $\underline{0.266{\scriptstyle\pm0.005}}$ & $0.294{\scriptstyle\pm0.001}$ & $0.190{\scriptstyle\pm0.004}$ & $\underline{0.187{\scriptstyle\pm0.009}}$ & $\underline{0.186{\scriptstyle\pm0.008}}$ & $0.154{\scriptstyle\pm0.001}$ & $0.138{\scriptstyle\pm0.004}$ \\
\midrule
\textbf{MWM (ours)} & $\mathbf{0.606{\scriptstyle\pm0.007}}$ & $\mathbf{0.575{\scriptstyle\pm0.035}}$ & $\mathbf{0.269{\scriptstyle\pm0.006}}$ & $\mathbf{0.318{\scriptstyle\pm0.004}}$ & $\mathbf{0.200{\scriptstyle\pm0.005}}$ & $\mathbf{0.202{\scriptstyle\pm0.007}}$ & $\mathbf{0.187{\scriptstyle\pm0.005}}$ & $\mathbf{0.165{\scriptstyle\pm0.002}}$ & $\mathbf{0.158{\scriptstyle\pm0.004}}$ \\
\bottomrule
\end{tabular}}
\end{table}

\noindent\textbf{Results.}
In Table~\ref{tab:cls-lp}, \method{} improves task-averaged balanced accuracy by \resultClsGainPct{}\% relative to CBraMod, the strongest baseline by this measure.

\subsection{Lateralization Analysis}
\label{sec:laterality-analysis}

We examine hand-motion readouts and motor-imagery topographies of the bilateral and lateral representations. The control, \emph{w/o FA}, removes the frame-averaging (FA) wrapper, leaving a plain Transformer. Protocols and additional analyses are provided in Appendix~\ref{app:laterality-analysis}.

\begin{figure}[t]
\centering
\includegraphics[width=\linewidth]{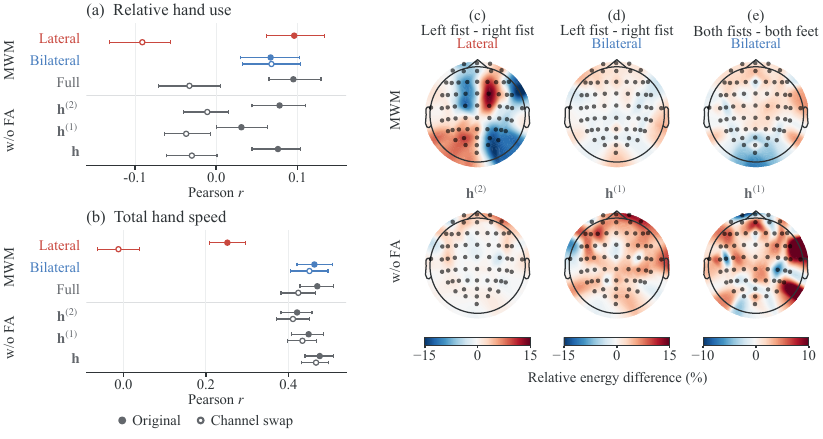}
\caption{\textbf{Lateralization analysis.} \textbf{(a,b)} Hand-motion correlations with 95\% confidence intervals estimated by bootstrap resampling, using participants as the resampling unit. \textbf{(c--e)} Participant-averaged motor-imagery feature-energy contrasts, normalized by channel-wise mean energy; left--right contrasts share a color scale. Control features $\mathbf{h}^{(1)}$ and $\mathbf{h}^{(2)}$ are equal-sized halves of $\mathbf{h}$.}
\label{fig:laterality-analysis}
\end{figure}

\noindent\textbf{Hand-motion readouts.}
We fit ridge readouts to frozen EgoBrain representations for relative hand use and total hand speed, using targets extracted from WiLoR~\citep{potamias2025wilor} hand tracks in synchronized video. In Figure~\ref{fig:laterality-analysis}(a,b), exchanging homologous raw EEG channels under the same readout reverses the lateral correlation for relative hand use from $0.096$ to $-0.091$, versus $0.078$ to $-0.011$ for the corresponding control block. The bilateral total-speed correlation changes little ($0.463$ to $0.451$). Both blocks encode both targets, but their contrasting responses support LAT's design of invariant bilateral and equivariant lateral readouts.

\noindent\textbf{Motor-imagery topographies.}
Figure~\ref{fig:laterality-analysis}(c--e) visualizes channel-wise feature-energy contrasts on held-out PhysioNet-MI data~\citep{schalk2009eegmmidb}. The left--right fist contrast is predominantly mirror-antisymmetric in the lateral block (fraction $0.811$), compared with $0.152$ for the bilateral block and $0.329$ for the corresponding control features. The bilateral both-fists versus both-feet contrast is predominantly symmetric. These patterns are consistent with LAT's different treatment of shared and side-specific activity.

\subsection{Ablation Studies}
\label{sec:ablations}

Table~\ref{tab:ablation} compares the full model with variants without frame averaging (w/o FA), prediction, or alignment. Removing any of these components reduces average transfer, supporting the combined use of hemispheric structure, predictive supervision, and cross-modal alignment.

\begin{table}[t]
\centering
\tiny
\setlength{\tabcolsep}{4pt}
\caption{Ablations of frame averaging (FA), prediction, and alignment. Balanced accuracy for classification and Pearson correlation for regression; mean $\pm$ sample s.d. over three seeds. The last row averages per-task relative changes with respect to MWM over the listed tasks.}
\label{tab:ablation}
\begin{tabular}{lcccc}
\toprule
Task & \textbf{MWM (ours)} & w/o FA & w/o prediction & w/o alignment \\
\midrule
\multicolumn{5}{l}{\emph{Classification}} \\
PhysioNet-MI & $\mathbf{0.606{\scriptstyle\pm0.007}}$ & $\mathbf{0.606{\scriptstyle\pm0.007}}$ & $\underline{0.600{\scriptstyle\pm0.009}}$ & $0.596{\scriptstyle\pm0.004}$ \\
Schwarz & $\mathbf{0.575{\scriptstyle\pm0.035}}$ & $\underline{0.570{\scriptstyle\pm0.015}}$ & $0.545{\scriptstyle\pm0.027}$ & $0.549{\scriptstyle\pm0.009}$ \\
Kaya5F & $\underline{0.269{\scriptstyle\pm0.006}}$ & $0.261{\scriptstyle\pm0.004}$ & $0.266{\scriptstyle\pm0.004}$ & $\mathbf{0.278{\scriptstyle\pm0.004}}$ \\
FineMI & $\underline{0.200{\scriptstyle\pm0.005}}$ & $\mathbf{0.201{\scriptstyle\pm0.003}}$ & $0.198{\scriptstyle\pm0.008}$ & $0.174{\scriptstyle\pm0.004}$ \\
Ofner2017MI & $\underline{0.165{\scriptstyle\pm0.002}}$ & $0.161{\scriptstyle\pm0.009}$ & $0.160{\scriptstyle\pm0.007}$ & $\mathbf{0.174{\scriptstyle\pm0.003}}$ \\
Jeong2020 & $\mathbf{0.158{\scriptstyle\pm0.004}}$ & $\mathbf{0.158{\scriptstyle\pm0.004}}$ & $\underline{0.149{\scriptstyle\pm0.006}}$ & $0.148{\scriptstyle\pm0.001}$ \\
\midrule
\multicolumn{5}{l}{\emph{Regression}} \\
ForceGameReg & $\underline{0.165{\scriptstyle\pm0.010}}$ & $0.164{\scriptstyle\pm0.007}$ & $\mathbf{0.171{\scriptstyle\pm0.014}}$ & $0.107{\scriptstyle\pm0.010}$ \\
SrisReachME & $\mathbf{0.241{\scriptstyle\pm0.002}}$ & $0.235{\scriptstyle\pm0.007}$ & $\underline{0.240{\scriptstyle\pm0.003}}$ & $0.166{\scriptstyle\pm0.000}$ \\
OfnerApertureME & $\underline{0.233{\scriptstyle\pm0.018}}$ & $\mathbf{0.236{\scriptstyle\pm0.021}}$ & $0.228{\scriptstyle\pm0.012}$ & $0.056{\scriptstyle\pm0.012}$ \\
WAYGALReg & $\mathbf{0.433{\scriptstyle\pm0.002}}$ & $\underline{0.391{\scriptstyle\pm0.003}}$ & $0.387{\scriptstyle\pm0.004}$ & $0.311{\scriptstyle\pm0.002}$ \\
BiomechMag & $\underline{0.157{\scriptstyle\pm0.002}}$ & $0.152{\scriptstyle\pm0.002}$ & $\mathbf{0.158{\scriptstyle\pm0.000}}$ & $0.144{\scriptstyle\pm0.001}$ \\
ForenzoFutVec & $\mathbf{0.127{\scriptstyle\pm0.001}}$ & $0.118{\scriptstyle\pm0.004}$ & $\underline{0.121{\scriptstyle\pm0.003}}$ & $0.108{\scriptstyle\pm0.002}$ \\
\midrule
Avg.\ rel.\ change & -- & $-2.3\%$ & $-2.6\%$ & $-17.6\%$ \\
\bottomrule
\end{tabular}
\end{table}

\subsection{Transfer World Models for Real-Action Conditioning}
\label{sec:world-action-probe}

We evaluate whether the pretrained world models transfer to future prediction conditioned on real actions. On EgoBrain~\citep{lin2026egobrain}, synchronized IMU measurements serve as a proxy for real actions in this evaluation. We train the same lightweight IMU-adapter architecture to map these measurements to each model's frozen action interface and predict frozen DINOv2~\citep{oquab2024dinov2} features $0.2$\,s ahead under within-recording and held-out-subject splits.

\noindent\textbf{Baselines and Evaluation.}
We compare \method{} with a predictor pretrained using inverse-dynamics latent actions~\citep{garrido2026lawm}, an action-free predictor pretrained with the identical recipe but without EEG, a randomly initialized predictor, and a copy-frame baseline that outputs the current frame. All models use the same inputs, targets, predictor architecture, IMU-adapter architecture, and action interface. Unlike the checkpoint used in the previous sections, all pretrained predictors are pretrained separately for each split using only its training portion. We report L1 error, cosine similarity, and the increase in error after shuffling the IMU inputs ($\Delta_{\mathrm{act}}$), with standard deviations from a subject-level bootstrap. Further details are provided in Appendix~\ref{app:world-model-probe}.

\begin{table}[ht]
\centering
\small
\setlength{\tabcolsep}{3pt}
\caption{Evaluation of real-action conditioning in held-out next-step prediction, with IMU measurements as the action proxy. All pretrained models use the same predictor and action interface. Copy frame outputs the current frame; the action-free predictor is pretrained with the identical recipe but without EEG. Values are mean $\pm$ standard deviation from a subject-level bootstrap; 95\% intervals are given in Appendix~\ref{app:world-model-probe}.}
\label{tab:wm-probe}
\resizebox{\textwidth}{!}{%
\begin{tabular}{lcccccc}
\toprule
& \multicolumn{3}{c}{Within recording}
& \multicolumn{3}{c}{Held-out subjects} \\
\cmidrule(lr){2-4}\cmidrule(lr){5-7}
Model & L1$\downarrow$ & Cos.$\uparrow$ & $\Delta_{\mathrm{act}}\uparrow$
& L1$\downarrow$ & Cos.$\uparrow$ & $\Delta_{\mathrm{act}}\uparrow$ \\
\midrule
Random init. & $0.7334{\scriptstyle\pm0.0165}$ & $0.8334{\scriptstyle\pm0.0078}$ & $+0.0000{\scriptstyle\pm0.0000}$ & $0.7224{\scriptstyle\pm0.0116}$ & $0.8407{\scriptstyle\pm0.0054}$ & $+0.0000{\scriptstyle\pm0.0000}$ \\
Copy frame & $0.5036{\scriptstyle\pm0.0234}$ & $0.8783{\scriptstyle\pm0.0082}$ & -- & $0.4961{\scriptstyle\pm0.0169}$ & $0.8859{\scriptstyle\pm0.0056}$ & -- \\
Action-free & $\underline{0.4754{\scriptstyle\pm0.0209}}$ & $\underline{0.8921{\scriptstyle\pm0.0070}}$ & $\underline{+0.0044{\scriptstyle\pm0.0005}}$ & $\underline{0.4753{\scriptstyle\pm0.0116}}$ & $0.8966{\scriptstyle\pm0.0041}$ & $\underline{+0.0021{\scriptstyle\pm0.0003}}$ \\
Latent action & $0.4787{\scriptstyle\pm0.0209}$ & $0.8909{\scriptstyle\pm0.0071}$ & $+0.0026{\scriptstyle\pm0.0003}$ & $0.4764{\scriptstyle\pm0.0106}$ & $\underline{0.8968{\scriptstyle\pm0.0042}}$ & $+0.0015{\scriptstyle\pm0.0003}$ \\
\textbf{\method{} (ours)} & $\mathbf{0.4694{\scriptstyle\pm0.0206}}$ & $\mathbf{0.8938{\scriptstyle\pm0.0069}}$ & $\mathbf{+0.0085{\scriptstyle\pm0.0006}}$ & $\mathbf{0.4675{\scriptstyle\pm0.0130}}$ & $\mathbf{0.8991{\scriptstyle\pm0.0044}}$ & $\mathbf{+0.0071{\scriptstyle\pm0.0005}}$ \\
\bottomrule
\end{tabular}}

\end{table}

\noindent\textbf{Results.}
Table~\ref{tab:wm-probe} shows that \method{} achieves the lowest L1 error and the largest action gap on both splits. Paired by subject, \method{} reduces L1 relative to the action-free predictor by $0.0060\pm0.0010$ within recordings and $0.0078\pm0.0026$ on held-out subjects. Its action gap is \resultWMWithinGapX{}$\times$ and \resultWMSubjectGapX{}$\times$ that of the latent-action model and about 2--3$\times$ that of the action-free predictor, indicating that the gain comes from action conditioning rather than from frame prediction alone. All pretrained predictors also improve over the copy-frame baseline. The randomly initialized predictor has a zero action gap, whereas \method{} remains sensitive to matched IMU inputs on both splits.

\section{Conclusion}

We introduced \method{}, which uses EEG as the latent action of a world model to learn motor representations without explicit action annotations. The Lateralization-Aware Transformer incorporates hemispheric correspondence through frame averaging. Relative to the strongest baseline in each task family, the mind model improves task-averaged balanced accuracy by \resultClsGainPct{}\% and Pearson correlation by \resultRegGainPct{}\%, while the world model supports future prediction conditioned on real actions. These results support synchronized neural and visual experience as supervision for motor decoding and world modeling. 

\noindent\textbf{Limitations.} Our world-model evaluation tests action-conditioned prediction error; establishing end-to-end evaluation, such as robotic planning, remains future work.

\section*{AI Use Statement}

We used generative AI tools solely to edit and polish the manuscript text, edit figures, and adjust the layout. All AI-assisted modifications were reviewed by the authors to ensure that the revised text preserved the intended technical meaning and that figure revisions preserved the underlying data and scientific content. We take full responsibility for the final content of this work, including all text, claims, and artifacts produced with the assistance of generative AI.

\section*{Ethics Statement}

This work studies motor EEG representation learning and action-conditioned visual prediction using the existing datasets described in Appendix~\ref{app:dataset-details}. We do not anticipate additional ethical concerns specific to the methods and experiments presented here.

\section*{Reproducibility Statement}

The Method section describes the model architecture and training objectives. Appendix~\ref{app:dataset-details} documents the pretraining corpus and all 16 downstream tasks, including preprocessing and data splits. Appendix~\ref{app:implementation-details} provides model configurations, optimization settings, and linear-probe protocols. Additional protocols and metrics for world-model evaluation and lateralization analysis are provided in Appendices~\ref{app:world-model-probe} and~\ref{app:laterality-analysis}, respectively. To facilitate reproduction, anonymized code is provided in the supplementary material.

\bibliography{iclr2027_conference}
\bibliographystyle{iclr2027_conference}

\clearpage
\appendix
\section{Dataset Details}
\label{app:dataset-details}

We describe the pretraining corpus and the sixteen downstream tasks reported in Tables~\ref{tab:reg-pooled-lp} and~\ref{tab:cls-lp}: seven regression tasks and nine classification tasks. To evaluate transfer to unseen participants, we use participant-disjoint train, validation, and test sets, shared by all models within each task. Participant assignments are specified below. All models receive the same task-specific one-second EEG interval sampled at 200\,Hz, with model-specific channel interfaces. EEG values in $\mu$V are divided by 100 before model input.

\subsection{Pretraining Corpus}

\noindent\textbf{EgoBrain}
EgoBrain~\citep{lin2026egobrain} records EEG, egocentric video, and wearable IMU from 40 participants performing everyday activities. We use the 32 EEG channels and egocentric video from all participants for pretraining. From each 114-channel recording, we retain the fixed set of 32 EEG channels. We then apply a $0.3$--$75$\,Hz band-pass filter and a 50\,Hz notch filter, resample the signal from 256 to 200\,Hz, convert it to $\mu$V, and clip it to $\pm300\,\mu$V. This preprocessing yields 56,950 non-overlapping four-second clips, totaling 63.28 hours. Video frames are resized to a short side of 256 pixels and center-cropped to $224\times224$. Each pretraining example contains a one-second EEG interval ending at an anchor frame, the anchor frame itself, and the target frame $0.2$\,s later. All participants are used for pretraining without a validation split, except in the world-model evaluation, where pretraining uses only the training portion of each split. The 4 arm- and wrist-mounted IMU units are not used for pretraining; they are used only for the world-model evaluation in Appendix~\ref{app:world-model-probe}.

\subsection{Continuous Motor Regression}

Except for ForceGameReg, the regression tasks use one-second EEG windows with a $0.2$\,s stride and predict task-specific force or movement quantities during the interval from $0.2$ to $0.4$\,s after each window. We standardize every target dimension using statistics estimated from up to 8,000 evenly spaced examples from the training split only.

\noindent\textbf{ForceGameReg}
ForceGameReg uses the Force Game dataset~\citep{li2026forcegame}. The study recruited 31 three-person groups; one non-compliant participant was excluded, leaving 64-channel EEG from 92 participants. The released 500\,Hz epochs were filtered at 1--40\,Hz by the dataset authors. We use EEG from $0.8$ to $1.8$\,s of each trial and predict the mean plateau force from $2$ to $4$\,s. We convert the EEG to $\mu$V, resample it to 200\,Hz, apply common-average referencing, and reject examples with $\max |\mathrm{EEG}|>800\,\mu$V. Groups, rather than individual participants, are assigned to splits so that participants sharing a trial remain together. The train, validation, and test splits contain 21,040, 2,681, and 3,556 examples, respectively.

\noindent\textbf{SrisReachME}
SrisReachME is derived from a discrete center-out reaching study~\citep{srisrisawang2024simultaneous}. We use 60 EEG channels from 19 participants, apply common-average referencing and a zero-phase 0.5\,Hz high-pass filter, and resample the recordings from 500 to 200\,Hz. We reject windows with $\max |\mathrm{EEG}|>800\,\mu$V. The target is the mean of the instantaneous planar reaching-speed magnitudes over the prediction interval. We use 13 participants for training, participants \{2, 8, 14\} for validation, and participants \{3, 9, 15\} for testing, resulting in 511,120, 118,314, and 115,879 examples.

\noindent\textbf{OfnerApertureME and OfnerKinME}
Both tasks use the same 61-channel motor-execution recordings from 15 participants~\citep{ofner2017upper}. We interpolate missing EEG values, apply common-average referencing and a zero-phase 0.5\,Hz high-pass filter, resample from 512 to 200\,Hz, and construct continuous windows that include inter-trial rest. We remove windows with $\max |\mathrm{EEG}|>800\,\mu$V. For OfnerApertureME, we standardize the 11--13 valid CyberGlove sensors within each participant, low-pass filter them at 1\,Hz, and fit the first principal component. The target is the mean absolute rate of change of this component. For OfnerKinME, we derive velocity from three-dimensional hand position sampled at 64\,Hz and predict the mean of the instantaneous three-dimensional speed magnitudes. Participants \{1, 4--13\}, \{2, 14\}, and \{3, 15\} form the train, validation, and test splits for both tasks, containing 175,950, 32,327, and 31,577 examples.

\noindent\textbf{WAYGALReg}
WAYGALReg uses the WAY-EEG-GAL grasp-and-lift recordings from 12 participants with 32 EEG channels at 500\,Hz~\citep{luciw2014multichannel}. We use the released per-lift EEG segments, apply a fourth-order zero-phase Butterworth high-pass filter at 0.5\,Hz within each lift, and form sliding windows within lifts. Each window is resampled to 200\,Hz and common-average referenced; windows with $\max |\mathrm{EEG}|>800\,\mu$V are removed. The two targets are mean absolute grip force and mean signed load force over the future interval. The train, validation, and test splits contain eight, two, and two participants: \{2--6, 10--12\}, \{7, 8\}, and \{1, 9\}, respectively.

\noindent\textbf{BiomechMag}
BiomechMag records wrist movements from 45 participants~\citep{perezblanco2026wrist} using eight central electrodes: C3, C1, Cz, C2, C4, CP3, CPz, and CP4. We use runs 3--9, apply common-average referencing, and resample the continuous recordings from 512 to 200\,Hz. We remove windows with $\max |\mathrm{EEG}|>800\,\mu$V. The target is the magnitude of the mean two-dimensional wrist-velocity vector over the prediction interval. A participant-level split based on participant index gives 27, 9, and 9 participants, corresponding to 369,642, 125,663, and 125,654 train, validation, and test examples.

\noindent\textbf{ForenzoFutVec}
ForenzoFutVec uses the same 28-participant continuous-pursuit motor-imagery BCI recordings as Forenzo2024~\citep{forenzo2024continuous}, with 62 EEG channels sampled at 1,000\,Hz. From each trial start, we construct one-second windows with a $0.2$\,s stride, resample each window to 200\,Hz, apply common-average referencing, and reject windows with $\max |\mathrm{EEG}|>800\,\mu$V. We limit each participant to approximately 20,000 windows, retaining the complete run in which this threshold is reached. The two outputs are the mean horizontal and vertical cursor velocities over the future interval, in workspace widths per second. These are closed-loop outputs of the online BCI decoder. The split uses S01--S12 and S21--S28 for training, S13--S16 for validation, and S17--S20 for testing, giving 20, 4, and 4 participants. This task and Forenzo2024 use distinct held-out participant sets.

\subsection{Motor Classification}

Each classification task uses a one-second EEG window at the task-specific time relative to the cue or movement onset specified below. All splits are participant-disjoint.

\noindent\textbf{PhysioNet-MI}
PhysioNet-MI~\citep{schalk2009eegmmidb} contains 64-channel motor-imagery EEG from 109 participants. We use the runs for left fist, right fist, both fists, and both feet, giving four classes. The EEG is common-average referenced, high-pass filtered at 0.3\,Hz, notch filtered at 60\,Hz, and resampled to 200\,Hz. We extract the four seconds following each cue and provide the first second to the model. The train, validation, and test splits contain 6,300, 1,734, and 1,803 examples from 70, 19, and 20 participants, respectively.

\noindent\textbf{Schwarz}
Schwarz uses the gel-electrode cohort of the BNCI 2020-001 reach-and-grasp dataset~\citep{schwarz2020analyzing}: 15 participants and 58 EEG channels at 256\,Hz. We evaluate pre-movement decoding of palmar versus lateral grasp from the $[-2,-1)$\,s interval relative to movement onset, during instructed target fixation; rest trials are excluded. We apply common-average referencing and resample to 200\,Hz. The train, validation, and test splits contain eleven, two, and two participants: \{1, 6--15\}, \{2, 3\}, and \{4, 5\}, respectively.

\noindent\textbf{Kaya5F}
Kaya5F uses the five-finger motor-imagery paradigm of \citet{kaya2018large}, comprising 19 sessions from eight participants. We retain 19 EEG channels after dropping A1, A2, and X5, map legacy temporal-electrode names to their modern equivalents, and apply common-average referencing. We take the first second from each marker-defined segment. Recordings at 1,000\,Hz are resampled to 200\,Hz; the other recordings are already at 200\,Hz. The five classes correspond to the thumb, index, middle, ring, and little fingers. The train, validation, and test splits contain four, two, and two participants: \{A, B, H, I\}, \{C, E\}, and \{F, G\}, respectively.

\noindent\textbf{Forenzo2024}
Forenzo2024 uses continuous-pursuit motor-imagery BCI recordings from 28 participants with 62 EEG channels sampled at 1,000\,Hz~\citep{forenzo2024continuous}. Acquisition included a 0.1--200\,Hz band-pass filter and a 60\,Hz notch filter. We extract non-overlapping one-second windows from each trial start, resample each window to 200\,Hz, apply common-average referencing, and reject windows with $\max |\mathrm{EEG}|>800\,\mu$V. Labels encode the dominant axis and sign of the mean target-to-cursor displacement within the window, giving right, up, left, and down classes. The split uses S01--S08 and S17--S28 for training, S09--S12 for validation, and S13--S16 for testing, giving 20, four, and four participants. The recordings combine the Main and Transfer-Learning study cohorts.

\noindent\textbf{FineMI}
FineMI~\citep{yi2025multimodal} contains eight classes of unilateral upper-limb motor imagery from 18 participants and 62 EEG channels. The released four-second epochs were common-average referenced and filtered at 4--40\,Hz by the dataset authors. Following the benchmark preprocessing, we convert the signals to $\mu$V, apply common-average referencing, resample from 250 to 200\,Hz, and use the first second of each epoch. The train, validation, and test splits contain 3,840, 960, and 960 examples from 12, 3, and 3 participants, respectively.

\noindent\textbf{Ofner2017ME and Ofner2017MI}
Ofner2017~\citep{ofner2017upper} provides matched motor-execution and motor-imagery sessions from 15 participants with 61 EEG channels. Both tasks classify elbow flexion, elbow extension, forearm supination, forearm pronation, hand closing, hand opening, and rest. We apply common-average referencing, high-pass filter at 0.5\,Hz, resample from 512 to 200\,Hz, and reject trials with $\max |\mathrm{EEG}|>800\,\mu$V. The model receives the first second of the three-second interval following the cue. Participants \{1, 4--13\}, \{2, 14\}, and \{3, 15\} form the train, validation, and test splits. These splits contain 4,452, 839, and 773 examples for motor execution and 4,601, 839, and 838 examples for motor imagery.

\noindent\textbf{Weibo2014}
Weibo2014~\citep{yi2014evaluation} contains six classes of simple or compound limb motor imagery and one rest class, recorded from 10 participants with 60 EEG channels. We evaluate the last second of motor imagery, $[6,7)$\,s relative to trial onset. We apply common-average referencing. The train, validation, and test splits contain 3,294, 1,118, and 1,118 examples from 6, 2, and 2 participants, respectively.

\noindent\textbf{Jeong2020}
Jeong2020 uses the motor-imagery recordings of participants 1--15 from \citet{jeong2020multimodal}. After removing EOG and EMG channels, we retain 60 EEG channels, apply common-average referencing, and resample from 1,000 to 200\,Hz. The eleven classes comprise six reaching directions, three grasp types, and two wrist-twisting movements. The model receives the first second after the cue. Participants \{1, 4--13\}, \{2, 14\}, and \{3, 15\} form the train, validation, and test splits, containing 16,850, 3,350, and 3,300 examples.

\section{Implementation Details}
\label{app:implementation-details}

\noindent\textbf{Model configuration.}
The EEG encoder uses $0.2$\,s temporal patches, a token width of 40, and 12 Transformer blocks with eight attention heads and feed-forward width 800. The first ten blocks produce the representation used for downstream evaluation, while the final two serve as the reconstruction decoder during pretraining. A global channel and a global temporal token augment the token grid. For $C$ EEG channels, the one-second downstream representation has shape $(C+1)\times6\times40$. The visual encoder is DINOv2-Base with $224\times224$ inputs and a $16\times16$ grid of 768-dimensional patch features. The frame predictor has two Transformer layers, width 512, eight attention heads, feed-forward width 1,024, and dropout 0.1, with a 128-dimensional action interface.

\noindent\textbf{Pretraining.}
We train \method{} in FP32 using AdamW with a batch size of 128, a learning rate of $5\times10^{-4}$, weight decay of 0.05, and gradient clipping at 0.8. The learning rate follows an iteration-wise cosine schedule to $10^{-5}$ without warmup. All reported downstream \method{} evaluations use the same fixed pretrained checkpoint. For full-model pretraining, we set $\lambda_{\mathrm{align}}=\lambda_{\mathrm{pred}}=1$.

\noindent\textbf{Linear-probe evaluation.}
We freeze each backbone in evaluation mode and train only a linear prediction head. For \method{}, we discard the visual encoder and frame predictor, disable stochastic mirrors, and flatten the frozen EEG token grid before applying the head. For regression, we additionally apply a non-affine LayerNorm and standardize each target dimension using training-split statistics.

We train all probes with AdamW using a learning rate of $5\times10^{-5}$, weight decay of 0.01, a batch size of 64, an iteration-wise cosine schedule to $10^{-6}$, and gradient clipping at 1. Classification probes are trained for 80 epochs with cross-entropy and 0.1 label smoothing, while regression probes are trained for 10 epochs with MSE. We select checkpoints by validation balanced accuracy for classification and validation Pearson correlation for regression.

\section{World-Model Evaluation Details}
\label{app:world-model-probe}

\noindent\textbf{Protocol}
IMU measurements serve as a proxy for real actions in this evaluation. Each example contains frozen DINOv2 patch features for the current frame and the frame $0.2$\,s later, together with three consecutive $0.2$\,s IMU summaries from four arm- and wrist-mounted sensors. Relative to the current frame at time $t$ and the target frame at $t+0.2$\,s, the three summaries cover $[t-0.2,t)$, $[t,t+0.2)$, and $[t+0.2,t+0.4)$\,s; the middle window is the movement between the two frames, and the neighboring windows absorb the IMU--EEG/video clock residual of about $0.1$\,s. We freeze each frame predictor and train only an IMU adapter, implemented as a three-layer MLP with 512-dimensional hidden features and LayerNorm. Inputs are standardized on the training set, and the adapter is trained for 4,000 steps with AdamW, a batch size of 64, and a learning rate of $10^{-3}$. All models use the same inputs, targets, adapter architecture, and training anchors.

\noindent\textbf{Splits}
The within-recording split reserves the final eleventh of each recording for evaluation and separates it from training by a $300$\,s guard interval. The held-out-subject split evaluates on P0006, P0016, P0026, and P0036. For each split, \method{} and the latent-action model are pretrained only on the training portion.

\noindent\textbf{Comparisons}
All compared predictors share the same frame-predictor architecture and an identical $128$-dimensional action interface, controlling for architecture and action dimensionality. For pretraining, \method{} uses one second of EEG ending at the current frame. The latent-action model~\citep{garrido2026lawm} instead derives its action from the current and target frames with an inverse-dynamics encoder. The action-free predictor is pretrained with the identical recipe but receives no EEG. We also evaluate a randomly initialized predictor and a copy-frame baseline that outputs the current-frame features. 

\noindent\textbf{Metrics}
We report the mean L1 error and patch-wise cosine similarity between predicted and target features. To measure use of the action inputs, we fix the current frame, shuffle IMU inputs among held-out examples from the same subject, and compute
\begin{equation}
    \Delta_{\mathrm{act}}=E_{\mathrm{shuf}}-E_{\mathrm{true}},
\end{equation}
where $E_{\mathrm{true}}$ and $E_{\mathrm{shuf}}$ are the L1 errors under matched and shuffled action inputs. A positive action gap indicates that matched inputs improve prediction. As adjacent examples within a subject are correlated, we resample held-out subjects 10{,}000 times, shared by all models (paired), and report the standard deviation of the bootstrap estimates. The corresponding subject-level 95\% intervals ($\pm1.96$ standard deviations) of $\Delta_{\mathrm{act}}$ in Table~\ref{tab:wm-probe} are $+0.0085\pm0.0012$ for \method{}, $+0.0044\pm0.0009$ for the action-free predictor, and $+0.0026\pm0.0006$ for the latent-action model within recordings, and $+0.0071\pm0.0011$, $+0.0021\pm0.0007$, and $+0.0015\pm0.0006$ on held-out subjects; the intervals of \method{} do not overlap with those of the other models on either split.

\noindent\textbf{Conditioning only on movement up to the target frame}
Table~\ref{tab:wm-probe-pretarget} repeats the evaluation of Table~\ref{tab:wm-probe} with only the two IMU windows that end at the target frame, $[t-0.2,t)$ and $[t,t+0.2)$. \method{} is nearly unchanged and retains the lowest L1 error on both splits. Its action gap remains larger than that of the action-free predictor on both splits.

\begin{table}[h]
\centering
\small
\setlength{\tabcolsep}{3pt}
\caption{Real-action conditioning with IMU windows up to the target frame only. Values are mean $\pm$ standard deviation from a subject-level bootstrap, as in Table~\ref{tab:wm-probe}.}
\label{tab:wm-probe-pretarget}
\resizebox{\textwidth}{!}{%
\begin{tabular}{lcccccc}
\toprule
& \multicolumn{3}{c}{Within recording}
& \multicolumn{3}{c}{Held-out subjects} \\
\cmidrule(lr){2-4}\cmidrule(lr){5-7}
Model & L1$\downarrow$ & Cos.$\uparrow$ & $\Delta_{\mathrm{act}}\uparrow$
& L1$\downarrow$ & Cos.$\uparrow$ & $\Delta_{\mathrm{act}}\uparrow$ \\
\midrule
Random init. & $0.7334{\scriptstyle\pm0.0165}$ & $0.8334{\scriptstyle\pm0.0078}$ & $+0.0000{\scriptstyle\pm0.0000}$ & $0.7224{\scriptstyle\pm0.0116}$ & $0.8407{\scriptstyle\pm0.0054}$ & $+0.0000{\scriptstyle\pm0.0000}$ \\
Copy frame & $0.5036{\scriptstyle\pm0.0234}$ & $0.8783{\scriptstyle\pm0.0082}$ & -- & $0.4961{\scriptstyle\pm0.0169}$ & $0.8858{\scriptstyle\pm0.0056}$ & -- \\
Action-free & $\underline{0.4753{\scriptstyle\pm0.0209}}$ & $\underline{0.8921{\scriptstyle\pm0.0070}}$ & $\underline{+0.0044{\scriptstyle\pm0.0005}}$ & $0.4754{\scriptstyle\pm0.0116}$ & $0.8966{\scriptstyle\pm0.0041}$ & $\underline{+0.0021{\scriptstyle\pm0.0003}}$ \\
Latent action & $0.4771{\scriptstyle\pm0.0208}$ & $0.8914{\scriptstyle\pm0.0071}$ & $+0.0030{\scriptstyle\pm0.0004}$ & $\underline{0.4753{\scriptstyle\pm0.0107}}$ & $\underline{0.8969{\scriptstyle\pm0.0041}}$ & $+0.0016{\scriptstyle\pm0.0003}$ \\
\textbf{\method{} (ours)} & $\mathbf{0.4693{\scriptstyle\pm0.0206}}$ & $\mathbf{0.8937{\scriptstyle\pm0.0069}}$ & $\mathbf{+0.0080{\scriptstyle\pm0.0006}}$ & $\mathbf{0.4677{\scriptstyle\pm0.0131}}$ & $\mathbf{0.8989{\scriptstyle\pm0.0045}}$ & $\mathbf{+0.0068{\scriptstyle\pm0.0006}}$ \\
\bottomrule
\end{tabular}}
\end{table}

\section{Lateralization Analysis Details}
\label{app:laterality-analysis}

We compare one fixed pretrained \method{} model with the \emph{w/o FA} variant, which removes the frame-averaging (FA) wrapper, leaving a plain Transformer. Both encoders are frozen, and preprocessed EEG inputs are multiplied by $1/100$ before encoding.

For a token $\mathbf{h}\in\mathbb{R}^{D_e}$, we write $\mathbf{h}=[\mathbf{h}^{(1)};\mathbf{h}^{(2)}]$, where each block lies in $\mathbb{R}^{D_e/2}$. In \method{}, these blocks correspond to bilateral and lateral features, respectively; in the control model, they are arbitrary partitions of the feature dimension.

\subsection{Hand-Motion Readouts}

\noindent\textbf{Data and targets.}
We uniformly sample 300 one-second windows from each of the 40 EgoBrain participants~\citep{lin2026egobrain}, retaining the 32-channel montage at 200\,Hz. We obtain 3-D hand tracks from synchronized video using WiLoR~\citep{potamias2025wilor} and compute ego-motion-compensated left- and right-hand speeds. Each window contains five $0.2$\,s frame slots. We retain windows with at least three valid slots for each hand, yielding 9,319 windows, and average the valid speeds within each window to obtain $L$ and $R$. The two targets are relative hand use \mbox{$y_{\mathrm{side}}=\log(R+\epsilon)-\log(L+\epsilon)$} and total hand speed $y_{\mathrm{speed}}=\log(R+L+\epsilon)$, with $\epsilon=10^{-3}$. Both targets are centered within each participant. All participants' EEG and video were available during pretraining; the hand labels were used only for this readout analysis.

\noindent\textbf{Features and evaluation.}
We fit separate ridge regressions to the global token $\mathbf{h}$ and its two blocks $\mathbf{h}^{(1)}$ and $\mathbf{h}^{(2)}$. Features are centered within each participant and standardized using training-fold statistics. We use five participant-disjoint folds, selecting the ridge penalty from $\{0.1,1,10,100\}$ by nested participant-grouped cross-validation. We report Pearson correlation after pooling the held-out predictions from all folds. Confidence intervals use 1,000 participant-bootstrap samples, shared across model and readout comparisons.

\noindent\textbf{Channel-exchange diagnostic.}
For each fold, we apply the fitted readout to features obtained after exchanging all 14 homologous electrode pairs in the raw EEG, leaving four midline channels and the targets unchanged. This diagnostic acts on raw channels; the training operation $\mathcal{P}$ instead exchanges only the learned lateral feature component. The readouts use the global representation, which has no channel axis.

\subsection{Motor-Imagery Topographies}

\noindent\textbf{Feature-energy contrasts.}
We use the first second after the cue from the 1,803 PhysioNet-MI test trials of 20 participants~\citep{schalk2009eegmmidb}: 451 left-fist, 449 right-fist, 450 both-fists, and 453 both-feet trials. Inputs retain 64 channels at 200\,Hz with a 0.3\,Hz high-pass filter. No prediction head is fitted. For each trial and channel, we sum squared patch-token values over the $N$ temporal patches and $D_e/2$ features of the selected block. After centering within participant, we compute each participant's mean left-fist minus right-fist energy contrast and then average across participants with equal weight. We similarly contrast both-fists with both-feet imagery using $\mathbf{h}^{(1)}$.

\noindent\textbf{Spatial symmetry.}
Let $\mathbf{m}\in\mathbb{R}^{64}$ be the group-mean energy contrast before channel-wise normalization, and let $\Pi$ exchange the 27 homologous channel pairs while fixing the ten midline channels. We measure the fraction of mirror-antisymmetric energy over all channels as
\begin{equation}
    A(\mathbf{m})=
    \frac{\|\mathbf{m}-\Pi\mathbf{m}\|_2^2}
    {\|\mathbf{m}-\Pi\mathbf{m}\|_2^2+\|\mathbf{m}+\Pi\mathbf{m}\|_2^2}.
\end{equation}
Figure~\ref{fig:laterality-analysis}(c--e) displays channel-normalized maps, while Table~\ref{tab:physionet-symmetry} reports $A$ computed from the unnormalized contrasts.

\begin{table}[t]
\centering
\small
\caption{Mirror-antisymmetric fraction $A$ of the left-fist minus right-fist energy contrast.}
\label{tab:physionet-symmetry}
\begin{tabular}{llc}
\toprule
Model & Feature block & $A$ \\
\midrule
\method{} & Lateral & 0.811 \\
\method{} & Bilateral & 0.152 \\
w/o FA & $\mathbf{h}^{(2)}$ & 0.329 \\
w/o FA & $\mathbf{h}^{(1)}$ & 0.250 \\
\bottomrule
\end{tabular}
\end{table}

\end{document}